# A Rule-Based Approach To Prepositional Phrase Attachment Disambiguation

**Eric Brill**
Spoken Language Systems Group
Laboratory for Computer Science, M.I.T.
Cambridge, Ma. 02139 U.S.A
brill@goldilocks.lcs.mit.edu

**Philip Resnik***
Sun Microsystems Laboratories, Inc.
Chelmsford, MA 01824–4195 U.S.A
philip.resnik@east.sun.com

## Abstract

In this paper, we describe a new corpus-based approach to prepositional phrase attachment disambiguation, and present results comparing performance of this algorithm with other corpus-based approaches to this problem.

## Introduction

Prepositional phrase attachment disambiguation is a difficult problem. Take, for example, the sentence:

(1) Buy a car [$_{PP}$ with a steering wheel].

We would guess that the correct interpretation is that one should buy cars that come with steering wheels, and not that one should use a steering wheel as barter for purchasing a car. In this case, we are helped by our world knowledge about automobiles and automobile parts, and about typical methods of barter, which we can draw upon to correctly disambiguate the sentence. Beyond possibly needing such rich semantic or conceptual information, Altmann and Steedman [AS88] show that there are certain cases where a discourse model is needed to correctly disambiguate prepositional phrase attachment.

However, while there are certainly cases of ambiguity that seem to need some deep knowledge, either linguistic or conceptual, one might ask what sort of performance could be achieved by a system that uses somewhat superficial knowledge automatically extracted from a large corpus. Recent work has shown that this approach holds promise [HR91, HR93].

In this paper we describe a new rule-based approach to prepositional phrase attachment disambiguation. A set of simple rules is learned automatically to try to predict proper attachment based on any of a number of possible contextual cues.

*Parts of this work done at the Computer and Information Science Department, University of Pennsylvania were supported by by DARPA and AFOSR jointly under grant No. AFOSR-90-0066, and by ARO grant No. DAAL 03-89-C0031 PRI (first author) and by an IBM graduate fellowship (second author). This work was also supported at MIT by ARPA under Contract N00014-89-J-1332, monitored through the Office of Naval research (first author).

## Baseline

Hindle and Rooth [HR91, HR93] describe a corpus-based approach to disambiguating between prepositional phrase attachment to the main verb and to the object noun phrase (such as in the example sentence above). They first point out that simple attachment strategies such as right association [Kim73] and minimal attachment [Fra78] do not work well in practice (see [WFB90]). They then suggest using lexical preference, estimated from a large corpus of text, as a method of resolving attachment ambiguity, a technique they call "lexical association." From a large corpus of parsed text, they first find all noun phrase heads, and then record the verb (if any) that precedes the head, and the preposition (if any) that follows it, as well as some other syntactic information about the sentence. An algorithm is then specified to try to extract attachment information from this table of co-occurrences. For instance, a table entry is considered a definite instance of the prepositional phrase attaching to the noun if:

> The noun phrase occurs in a context where no verb could license the prepositional phrase, specifically if the noun phrase is in a subject or other pre-verbal position.

They specify seven different procedures for deciding whether a table entry is an instance of no attachment, sure noun attach, sure verb attach, or an ambiguous attach. Using these procedures, they are able to extract frequency information, counting the number of times a particular verb or noun appears with a particular preposition.

These frequencies serve as training data for the statistical model they use to predict correct attachment. To disambiguate sentence (1), they would compute the likelihood of the preposition *with* given the verb *buy*, and contrast that with the likelihood of that preposition given the noun *wheel*.

One problem with this approach is that it is limited in what relationships are examined to make an attachment decision. Simply extending Hindle and Rooth's model to allow for relationships such as that between the verb and the object of the preposition would result in too large a parameter space, given any realistic quantity of training data. Another problem of the method, shared by many statistical approaches, is that the model acquired during training is represented in a huge table of probabilities, precluding any straightforward analysis of its workings.

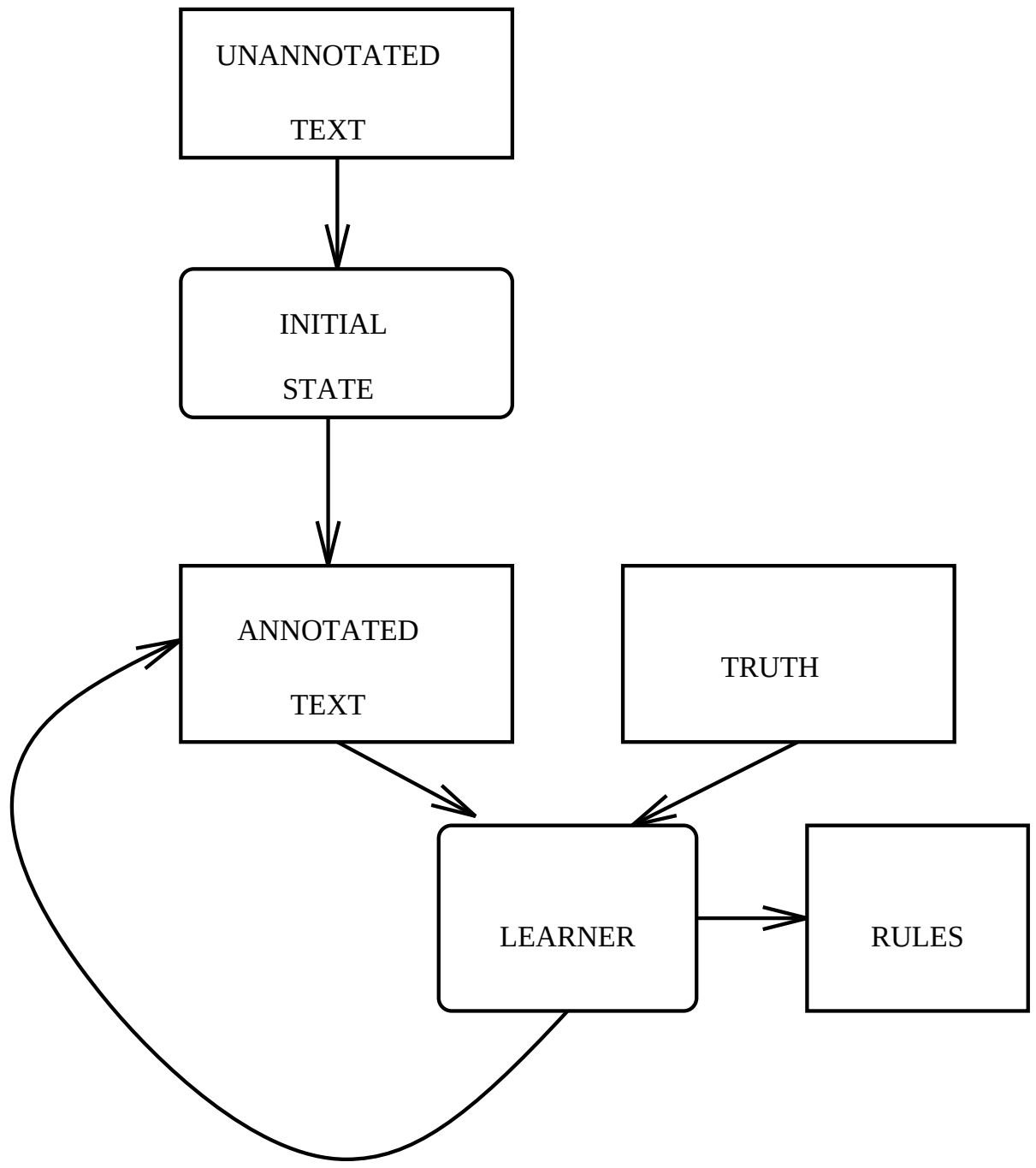

Figure 1: Transformation-Based Error-Driven Learning.

## Transformation-Based Error-Driven Learning

Transformation-based error-driven learning is a simple learning algorithm that has been applied to a number of natural language problems, including part of speech tagging and syntactic parsing [Bri92, Bri93a, Bri93b, Bri94]. Figure 1 illustrates the learning process. First, unannotated text is passed through the initial-state annotator. The initial-state annotator can range in complexity from quite trivial (e.g. assigning random structure) to quite sophisticated (e.g. assigning the output of a knowledge-based annotator that was created by hand). Once text has been passed through the initial-state annotator, it is then compared to the *truth*, as indicated in a manually annotated corpus, and transformations are learned that can be applied to the output of the initial state annotator to make it better resemble the *truth*.

So far, only a greedy search approach has been used: at each iteration of learning, the transformation is found whose application results in the greatest improvement; that transformation is then added to the ordered transformation list and the corpus is updated by applying the learned transformation. (See [RM94] for a detailed discussion of this algorithm in the context of machine learning issues.)

Once an ordered list of transformations is learned, new text can be annotated by first applying the initial state annotator to it and then applying each of the transformations, in order.

## Transformation-Based Prepositional Phrase Attachment

We will now show how transformation-based error-driven learning can be used to resolve prepositional phrase attachment ambiguity. The prepositional phrase attachment learner learns transformations from a corpus of 4-tuples of the form (v n1 p n2), where v is a verb, n1 is the head of its object noun phrase, p is the preposition, and n2 is the head of the noun phrase governed by the preposition (for example, *see/v the boy/n1 on/p the hill/n2*). For all sentences that conform to this pattern in the Penn Treebank Wall Street Journal corpus [MSM93], such a 4-tuple was formed, and each 4-tuple was paired with the attachment decision used in the Treebank parse.[1] There were 12,766 4-tuples in all, which were randomly split into 12,266 training samples and 500 test samples. In this experiment (as in [HR91, HR93]), the attachment choice for prepositional phrases was between the object noun and the matrix verb. In the initial state annotator, all prepositional phrases are attached to the object noun.[2] This is the attachment predicted by right association [Kim73].

The allowable transformations are described by the following templates:

- Change the attachment location from X to Y if:
  - n1 is W
  - n2 is W
  - v is W
  - p is W
  - n1 is W1 and n2 is W2
  - n1 is W1 and v is W2
  - ...

Here "from X to Y" can be either "from n1 to v" or "from v to n1," W (W1, W2, etc.) can be any word, and the ellipsis indicates that the complete set of transformations permits matching on *any* combination of values for v, n1, p, and n2, with the exception of patterns that specify values for all four. For example, one allowable transformation would be

[1] Patterns were extracted using `tgrep`, a tree-based grep program written by Rich Pito. The 4-tuples were extracted automatically, and mistakes were not manually pruned out.

[2] If it is the case that attaching to the verb would be a better start state in some corpora, this decision could be parameterized.

Change the attachment location from n1 to v
if p is "until".

Learning proceeds as follows. First, the training set is processed according to the start state annotator, in this case attaching all prepositional phrases low (attached to n1). Then, in essence, each possible transformation is scored by applying it to the corpus and computing the reduction (or increase) in error rate. In reality, the search is data driven, and so the vast majority of allowable transformations are not examined. The best-scoring transformation then becomes the first transformation in the learned list. It is applied to the training corpus, and learning continues on the modified corpus. This process is iterated until no rule can be found that reduces the error rate.

In the experiment, a total of 471 transformations were learned — Figure 3 shows the first twenty.[3] Initial accuracy on the test set is 64.0% when prepositional phrases are always attached to the object noun. After applying the transformations, accuracy increases to 80.8%. Figure 2 shows a plot of test-set accuracy as a function of the number of training instances. It is interesting to note that the accuracy curve has not yet reached a plateau, suggesting that more training data would lead to further improvements.

## Adding Word Class Information

In the above experiment, all transformations are triggered by words or groups of words, and it is surprising that good performance is achieved even in spite of the inevitable sparse data problems. There are a number of ways to address the sparse data problem. One of the obvious ways, mapping words to part of speech, seems unlikely to help. Instead, semantic class information is an attractive alternative.

We incorporated the idea of using semantic information in the following way. Using the WordNet noun hierarchy [Mil90], each noun in the training and test corpus was associated with a set containing the noun itself plus the name of every semantic class that noun appears in (if any).[4] The transformation template is modified so that in addition to asking if a noun matches some word W, it can also ask if it is a member of some class C.[5] This approach to data sparseness is similar to that of [Res93b, RH93], where a method is proposed for using WordNet in conjunction with a corpus to obtain class-based statistics. Our method here is much simpler, however, in that we are only using Boolean values to indicate whether a word can be a member of a class, rather than estimating a full set of joint probabilities involving classes.

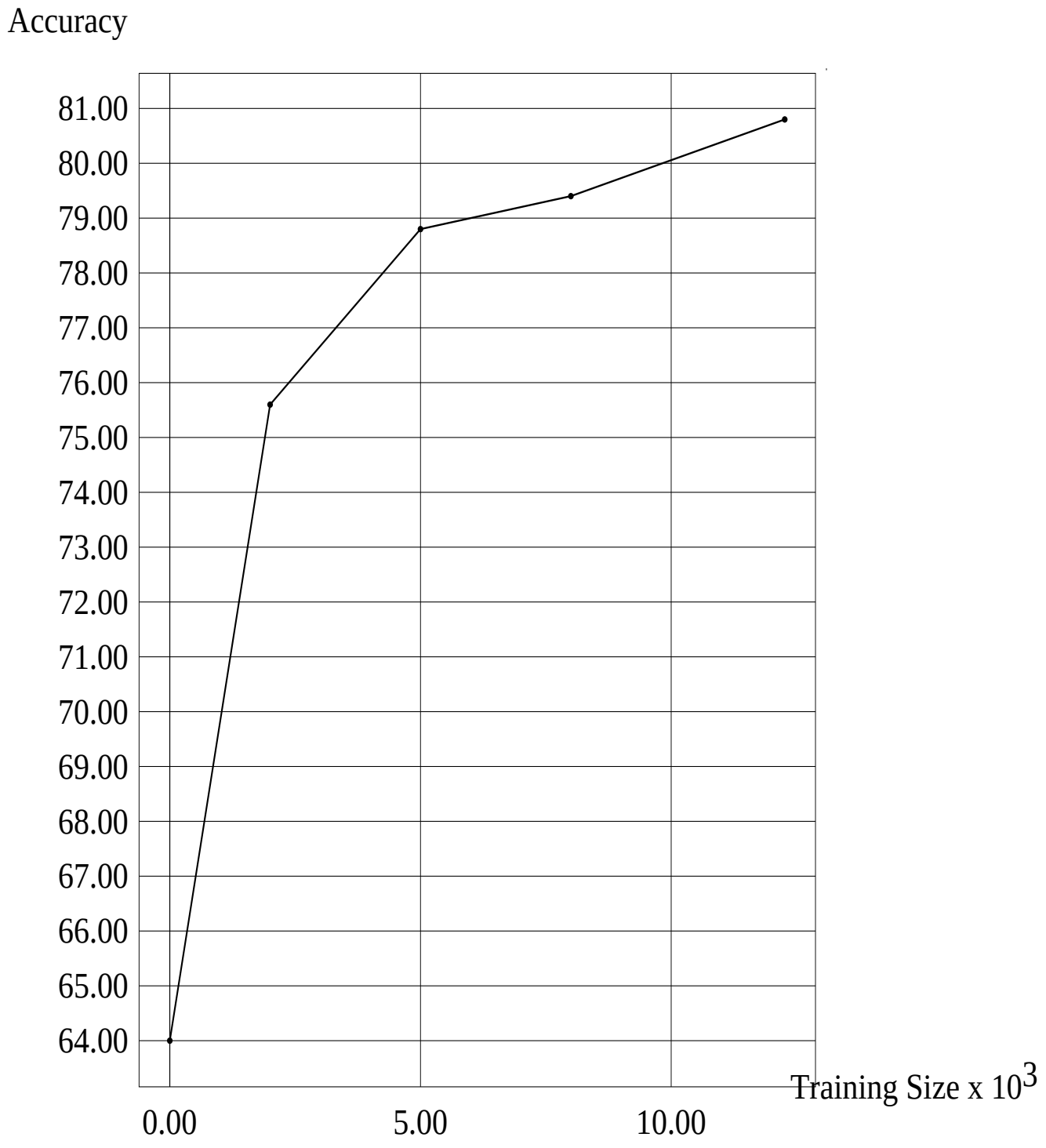

Figure 2: Accuracy as a function of training corpus size (no word class information).

[3]In transformation #8, word token `amount` appears because it was used as the head noun for noun phrases representing percentage amounts, e.g. "5%." The rule captures the very regular appearance in the Penn Treebank Wall Street Journal corpus of parses like *Sales for the year* [$_{VP}$ *rose* [$_{NP}$*5%*][$_{PP}$ *in fiscal 1988*]].

[4]Class names corresponded to unique "synonym set" identifiers within the WordNet noun database. A noun "appears in" a class if it falls within the hyponym (IS-A) tree below that class. In the experiments reported here we used WordNet version 1.2.

[5]For reasons of run-time efficiency, transformations making reference to the classes of both n1 and n2 were not permitted.

| # | Change Attachment Location From | To | Condition |
|---|---|---|---|
| 1 | N1 | V | P is *at* |
| 2 | N1 | V | P is *as* |
| 3 | N1 | V | P is *into* |
| 4 | N1 | V | P is *from* |
| 5 | N1 | V | P is *with* |
| 6 | N1 | V | N2 is *year* |
| 7 | N1 | V | P is *by* |
| 8 | N1 | V | P is *in* and N1 is *amount* |
| 9 | N1 | V | P is *through* |
| 10 | N1 | V | P is *during* |
| 11 | N1 | V | V is *put* |
| 12 | N1 | V | N2 is *month* |
| 13 | N1 | V | P is *under* |
| 14 | N1 | V | P is *after* |
| 15 | N1 | V | V is *have* and P is *in* |
| 16 | N1 | V | P is *without* |
| 17 | V | N1 | P is *of* |
| 18 | N1 | V | V is *buy* and P is *for* |
| 19 | N1 | V | P is *before* |
| 20 | N1 | V | V is *have* and P is *on* |

Figure 3: The first 20 transformations learned for prepositional phrase attachment.

Since the transformation-based approach with classes can generalize in a way that the approach without classes is unable to, we would expect fewer transformations to be necessary. Experimentally, this is indeed the case. In a second experiment, training and testing were carried out on the same samples as in the previous experiment, but this time using the extended transformation templates for word classes. A total of 266 transformations were learned. Applying these transformations to the test set resulted in an accuracy of 81.8%. In figure 4 we show the first 20 transformations learned using noun classes. Class descriptions are surrounded by square brackets.[6] The first transformation states that if N2 is a noun that describes time (i.e. is a member of WordNet class that includes the nouns "year," "month," "week," and others), then the prepositional phrase should be attached to the verb, since time is much more likely to modify a verb (e.g. *leave the meeting in an hour*) than a noun.

This experiment also demonstrates how any feature-based lexicon or word classification scheme can trivially be incorporated into the learner, by extending transformations to allow them to make reference to a word and any of its features.

## Evaluation against Other Algorithms

In [HR91, HR93], training is done on a superset of sentence types used in training the transformation-based learner. The transformation-based learner is trained on sentences containing v, n1 and p, whereas the algorithm described by Hindle and Rooth can also use sentences containing only v and p, or only n1 and p. In their paper, they train on over 200,000 sentences with prepositions from the Associated Press (AP) newswire, and they quote an accuracy of 78-80% on AP test data.

In order to compare the two approaches, we reimplemented the algorithm from [HR91] and tested it using the same training and test set used for the above experiments. Doing so resulted in an attachment accuracy of 70.4%. Next, the training set was expanded to include not only the cases of ambiguous attachment found in the parsed Wall Street Journal corpus, as before, but also all the unambiguous prepositional phrase attachments found in the corpus, as well (continuing to exclude the test set, of course). Accu-racy improved to 75.8%[7] using the larger training set, still significantly lower than accuracy obtained using the transformation-based approach. The technique described in [Res93b, RH93], which combined Hindle and Rooth's lexical association technique with a WordNet-based conceptual association measure, resulted in an accuracy of 76.0%, also lower than the results obtained using transformations.

[6] For expository purposes, the unique WordNet identifiers have been replaced by words that describe the content of the class.

| # | Change Attachment Location From | To | Condition |
|---|---|---|---|
| 1 | N1 | V | N2 is *[time]* |
| 2 | N1 | V | P is *at* |
| 3 | N1 | V | P is *as* |
| 4 | N1 | V | P is *into* |
| 5 | N1 | V | P is *from* |
| 6 | N1 | V | P is *with* |
| 7 | N1 | V | P is *of* |
| 8 | N1 | V | P is *in* and N1 is *[measure,quantity,amount]* |
| 9 | N1 | V | P is *by* and N2 is *[abstraction]* |
| 10 | N1 | V | P is *through* |
| 11 | N1 | V | P is *in* and N1 is *[group,grouping]* |
| 12 | V | N1 | V is *be* |
| 13 | N1 | V | V is *put* |
| 14 | N1 | V | P is *under* |
| 15 | N1 | V | P is *in* and N1 is *[written communication]* |
| 16 | N1 | V | P is *without* |
| 17 | N1 | V | P is *during* |
| 18 | N1 | V | P is *on* and N1 is *[thing]* |
| 19 | N1 | V | P is *after* |
| 20 | N1 | V | V is *buy* and P is *for* |

Figure 4: The first 20 transformations learned for prepositional phrase attachment, using noun classes.

| Method | Accuracy | # of Transforms |
|---|---|---|
| t-Scores | 70.4 - 75.8 | |
| Transformations | 80.8 | 471 |
| Transformations (no N2) | 79.2 | 418 |
| Transformations (classes) | 81.8 | 266 |

Figure 5: Comparing Results in PP Attachment.

Since Hindle and Rooth's approach does not make reference to n2, we re-ran the transformation-learner disallowing all transformations that make reference to n2. Doing so resulted in an accuracy of 79.2%. See figure 5 for a summary of results.

It is possible to compare the results described here with a somewhat similar approach developed independently by Ratnaparkhi and Roukos [RR94], since they also used training and test data drawn from the Penn Treebank's Wall Street Journal corpus. Instead of using manually constructed lexical classes, they use word classes arrived at via mutual information clustering in a training corpus [BDd+92], resulting in a representation in which each word is represented by a sequence of bits. As in the experiments here, their statistical model also makes use of a 4-tuple context (v, n1, p, n2), and can use the identities of the words, class information (for them, values of any of the class bits), or both kinds of information as contextual features — they describe a search process used to determine what subset of the available information will be used in the model. Given a choice of features, they train a probabilistic model for Pr(Site|context), and in testing choose Site = v or Site = n1 according to which has the higher conditional probability.

Ratnaparkhi and Roukos report an accuracy of 81.6% using both word and class information on Wall Street Journal text, using a training corpus twice as large as that used in our experiments. They also report that a decision tree model constructed using the same features and training data achieved performance of 77.7% on the same test set.

[7] The difference between these results and the result they quoted is likely due to a much larger training set used in their original experiments.

A number of other researchers have explored corpus-based approaches to prepositional phrase attachment disambiguation that make use of word classes. For example, Weischedel *et al.* [WAB+91] and Basili *et al.* [BPV91] both describe the use of manually constructed, domain-specific word classes together with corpus-based statistics in order to resolve prepositional phrase attachment ambiguity. Because these papers describe results obtained on different corpora, however, it is difficult to make a performance comparison.

## Conclusions

The transformation-based approach to resolving prepositional phrase disambiguation has a number of advantages over other approaches. In a direct comparison with lexical association, higher accuracy is achieved using words alone even though attachment information is captured in a relatively small number of simple, readable rules, as opposed to a large number of lexical co-occurrence probabilities.

In addition, we have shown how the transformation-based learner can easily be extended to incorporate word-class information. This resulted in a slight increase in performance, but more notably it resulted in a reduction by roughly half in the total number of transformation rules needed. And in contrast to approaches using class-based probabilistic models [BPV91, Res93c, WAB+91] or classes derived via statistical clustering methods [RR94], this technique produces a rule set that captures conceptual generalizations concisely and in human-readable form.

Furthermore, insofar as comparisons can be made among separate experiments using Wall Street Journal training and test data ([HR91], reimplemented as reported above; [Res93c, RH93]; [RR94]), the rule-based approach described here achieves better performance, using an algorithm that is conceptually quite simple and in practical terms extremely easy to implement.[8]

A more general point is that the transformation-based approach is easily adapted to situations in which some learning from a corpus is desirable, but hand-constructed prior knowledge is also available. Existing knowledge, such as structural strategies or even *a priori* lexical preferences, can be incorporated into the start state annotator, so that the learning algorithm begins with more refined input. And known exceptions can be handled transparently simply by adding additional rules to the set that is learned, using the same representation.

A disadvantage of the approach is that it requires supervised training — that is, a representative set of "true" cases from which to learn. However, this becomes less of a problem as annotated corpora become increasingly available, and suggests the combination of supervised and unsupervised methods as an interesting avenue for further research.

[8] Our code is being made publicly available. Contact the authors for information on how to obtain it.